\documentclass[sigconf]{acmart}

\AtBeginDocument{%
  }

\setcopyright{acmcopyright}
\copyrightyear{2018}
\acmYear{2018}
\acmDOI{XXXXXXX.XXXXXXX}

\acmConference[Conference acronym 'XX]{Make sure to enter the correct
  conference title from your rights confirmation emai}{June 03--05,
  2018}{Woodstock, NY}
\acmPrice{15.00}
\acmISBN{978-1-4503-XXXX-X/18/06}

\copyrightyear{2023}
\acmYear{2023}
\setcopyright{rightsretained}
\acmConference[CODS-COMAD 2023]{6th Joint International Conference on Data Science & Management of Data (10th ACM IKDD CODS and 28th COMAD)}{January 4--7, 2023}{Mumbai, India}
\acmBooktitle{6th Joint International Conference on Data Science \& Management of Data (10th ACM IKDD CODS and 28th COMAD) (CODS-COMAD 2023), January 4--7, 2023, Mumbai, India}\acmDOI{10.1145/3570991.3571004}
\acmISBN{978-1-4503-9797-1/23/01}
\begin{document}

\title{Deep Gaussian Processes for Air Quality Inference}

\author{Aadesh Desai*, Eshan Gujarathi*, Saagar Parikh*, Sachin Yadav*, Zeel Patel, Nipun Batra}
\thanks{*The authors contributed equally, and are undergraduate students.} 
\email{{desai.aadesh, eshan.rg, saagar.p, sachin.yadav, patel_zeel, nipun.batra}@iitgn.ac.in}
\affiliation{
  \institution{Indian Institute of Technology Gandhinagar}
  \city{Gandhinagar}
  \country{India}
  }

\renewcommand{\shortauthors}{Desai et al.}

\begin{abstract}

Air pollution kills around 7 million people annually, and approximately 2.4 billion people are exposed to hazardous air pollution. Accurate, fine-grained air quality (AQ) monitoring is essential to control and reduce pollution. However, AQ station deployment is sparse, and thus air quality inference for unmonitored locations is crucial. Conventional interpolation methods fail to learn the complex AQ phenomena. This work demonstrates that Deep Gaussian Process models (DGPs) are a promising model for the task of AQ inference. We implement Doubly Stochastic Variational Inference, a DGP algorithm, and show that it performs comparably to the state-of-the-art models.

\end{abstract}






\maketitle

\section{Introduction}



Air pollution causes fine particulate matter which results in strokes, heart diseases, lung cancer, and acute and chronic respiratory diseases
Accurate fine-grained monitoring of air quality can help mitigate air pollution. The air quality monitoring stations are expensive to install and maintain. In this paper, we study AQ inference to estimate AQ at unmonitored locations.

Prior works like \textbf{ADAIN}~\cite{cheng2018neural} have leveraged attention-based mechanisms to estimate AQ. But, the deep neural network cannot capture the notion of uncertainty. This limitation can be solved by using non-parametric probabilistic machine learning techniques such as \textbf{Gaussian Processes} (GPs) ~\cite{mackay1998introduction}. The ability of GPs to capture domain knowledge while predicting makes them advantageous over deep learning methods. Hence, GPs play a crucial role in certain applications where intuitive risk assessments are needed~\cite{sander_2021}. However, GPs are limited by issues such as scalability~\cite{liu2020gaussian} and the inability to extract hierarchical features~\cite{jakkala2021deep}.

Environmental phenomena exhibit non-stationarity (dynamic mean and covariance matrix) and complex inter-variable relationships. However, if we use stationary GPs, we need to parameterise the discontinuities, which becomes a problem if we do not know the number of discontinuities and do not wish to specify where they occur. In \textbf{Deep Gaussian Processes} (DGPs), we can obtain a non-parametric representation of discontinuities since the models are not restricted to Gaussian processes, and the covariance function can have non-Gaussian characteristics\footnote{http://inverseprobability.com/talks/notes/deep-gaussian-processes.html}. DGPs are multi-layer hierarchical generalisations of GPs, comparable to neural networks with numerous, indefinitely large hidden layers. DGPs are probabilistic and non-parametric, making them more generalizable, and capable of providing better-calibrated uncertainty estimates than other deep models.

Prior literature~\cite{damianou2013deep} has shown that DGPs have outperformed GPs on several known datasets, thus encouraging us to endeavour this approach for AQ inference. Further, prior work~\cite{DSVI} shows that their proposed method, Doubly Stochastic Variational Inference (DSVI) algorithm, a DGP model, works well in practice on different classification and regression tasks. For the regression task on the UCI Elevators Dataset, we found that DGP model outperforms the Gaussian Processes~\cite{patel2022accurate} model and other baselines. DSVI has also been proven~\cite{DSVI} to work well for other large-scale regression tasks, this motivated us to pursue DSVI for air quality inference on the Beijing Dataset. 

Our experiments show promise in that the air quality inference can be improved by combining the ardent power of deep learning with GPs in one expressive Bayesian learning model.

\section{Approach}

We implement the Doubly Stochastic Variational Inference (DSVI) model which incorporates sparsity, stochastically approximates the Bayesian model and returns a posterior using Variational Inference. Sparse variational inference employs inducing points to simplify the complex correlation between the DGP layers. While evaluation, the model takes random samples from univariate Gaussians. 

We use GPyTorch~\cite{gardner2018gpytorch} to build a DSVI Deep Gaussian model. We use the Squared Exponential (RBF) kernel and the \textit{VariationalELBO} loss function. The layers also have "skip connections" between them, similar to ResNet~\cite{he2016deep}.

\begin{table*}[]
\caption{Comparison of Doubly Stochastic Variational Inference DGP model with baselines on the Beijing Air Quality dataset.}
\begin{tabular}{l@{\hskip 0.25in}c@{\hskip 0.25in}c@{\hskip 0.25in}r@{\hskip 0.15in}}
\hline
\multicolumn{1}{c}{Model} & RMSE \(\downarrow\) & MAE \(\downarrow\) & R\(^2\) \(\uparrow\) \\ \hline
Random Forest~\cite{randomforest} & 28.136 & 16.178 & 0.851 \\
Inverse Distance Weighting (IDW) ~\cite{idw} & 48.281 & 33.888 & -0.215 \\
XGBoost (XGB)~\cite{xgb} & 35.526 & 24.866 & 0.695 \\
Support Vector Regressor (SVR)~\cite{svm} & 56.629 & 39.479 & -0.378 \\
K-Nearest Neighbours (KNN)~\cite{knn} & 40.819 & 26.108 & 0.639 \\
GP Model A\(\overline{\text{N}}\)CL~\cite{patel2022accurate} & 27.836 & 16.732 & 0.863 \\
GP Model A\(\overline{\text{N}}\)C\(\overline{\text{L}}\)~\cite{patel2022accurate} & \textbf{26.508} & \textbf{16.017} & \textbf{0.875} \\ \hline
Doubly Stochastic Variational Inference & 30.158 & 19.302 & 0.824 \\ \hline
\end{tabular}
\label{table:results}
\end{table*}
\section{Evaluation}

\subsection{Experimental Setup}

The dataset comprises of hourly PM\(_{2.5}\) (major pollutant) data from 36 air quality monitoring stations located in Beijing city along with meteorological data (temperature, weather, humidity, wind direction and pressure). The dataset is available for duration of a year from 1\(^\text{st}\) May 2014 to 30\(^\text{th}\) April 2015.

Following previous works~\cite{cheng2018neural, patel2022accurate} we use 3-fold cross validation. We perform hyperparameter tuning using grid search on \textit{inducing points} \(\in\) \{25, 40, 60, 100\}, \textit{number of samples} \(\in\) \{3, 5, 7\}, and \textit{learning rate} \(\in\) \{0.01, 0.05, 0.1\}.

We finally obtain the number of inducing points as 100 and number of samples as 7. We train our model for 500 epochs at a learning rate of 0.05. We use Root Mean Squared Error (RMSE), Mean Absolute Error (MAE) and the R\(^2\) score as the evaluation metrics for comparison.

\subsection{Results}
\begin{figure}[h!]
  \centering
  \includegraphics[width=0.9\linewidth]{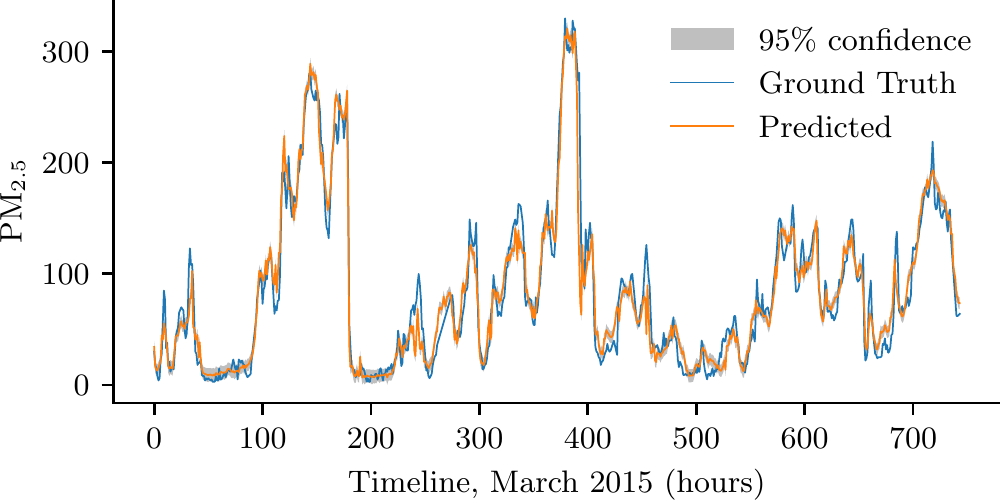}
  \caption{Predicted PM\(_{2.5}\) comparison with the ground truth values for DSVI model at station 1006}
  \label{fig:dsviplot}
\end{figure}

We have shown the experimental results in Table \ref{table:results}. We compare the Doubly Stochastic Variational Inference model with the state-of-the-art GPs, and other widely known baselines which have been used in previous literature. Our approach is comparable to the state-of-the-art. Figure \ref{fig:dsviplot} shows the predicted values of Doubly Stochastic Variational Inference model with the ground truth values at the location of station 1006. In the figure, the confidence band shows the deviation from the actual predicted value, which is one of the benefits of using DGPs over several baselines.

We identify the following areas where the model can be improved: 1) Figure \ref{fig:dsviplot} shows that the model is able to capture low-frequency signal correctly but is unable to predict the high-frequency signals accurately. 2) Also, we are not able to detect uncertainty band near the peaks.

\section{Future Work}
We now discuss the future work to overcome the limitations. We can improve the predictions by i) using combinations of kernels ii) designing kernels pertaining to specific features~\cite{patel2022accurate}. iii) using calibration to better capture the notion of uncertainty and iv) increasing the number of layers to further exploit the advantages of DGPs and tune the hyperparameters accordingly

\section{Conclusion}
Accurate AQ inference is an essential task for controlling air pollution. We show that DGPs are able to leverage domain insights such as: periodicity in time, inherent non-stationarity and the relative importance of hierarchical features. The Doubly Stochastic Variational Inference model has a simple kernel domain but still gives comparable results with state-of-the-art Gaussian Processes models. Hence, with this work, we encourage research to further explore Deep Gaussian Processes for AQ inference.

\bibliographystyle{ACM-Reference-Format}
\bibliography{sample-base}

\end{document}